\documentclass{article}

\usepackage{arxiv}

\usepackage{graphicx}

\usepackage[linesnumbered,ruled]{algorithm2e}
\usepackage{amsmath}
\usepackage{amsfonts}
\usepackage{multirow,changepage }

\usepackage{setspace}
\usepackage{url}

\usepackage{blindtext}
\usepackage{multicol}
\usepackage{color}

\usepackage{comment}
\usepackage{wrapfig}
\usepackage{graphicx}

\usepackage{booktabs} 

\usepackage{silence}
\usepackage{framed,enumitem,amssymb}
\newlist{todolist}{itemize}{2}
\setlist[todolist]{label=$\square$}
\usepackage{pifont}

\usepackage{listings}
\usepackage{threeparttable}
\usepackage{hyperref}
\usepackage{enumitem}
\usepackage{amsmath}
\usepackage{subcaption}

\usepackage[dvipsnames]{xcolor}
\usepackage{tcolorbox}
\tcbuselibrary{skins}

\usepackage{url}
\usepackage{xspace}
\usepackage[utf8]{inputenc}
\usepackage{cuted}

\usepackage[T1]{fontenc}
\usepackage{multicol}

\usepackage{diagbox}
\usepackage{multirow}

\usepackage[noadjust]{cite}

\definecolor{codegreen}{rgb}{0,0.6,0}
\definecolor{codegray}{rgb}{0.5,0.5,0.5}
\definecolor{codepurple}{rgb}{0.58,0,0.82}
\definecolor{backcolour}{rgb}{0.95,0.95,0.99}

\newif\iftodo
\todotrue

\let\OLDthebibliography\thebibliography
\renewcommand\thebibliography[1]{
  \OLDthebibliography{#1}
  \setlength{\parskip}{0pt}
  \setlength{\itemsep}{0pt plus 0.3ex}
}

\renewcommand{\textsf}{\texttt}

\usepackage{amsmath}

\usepackage{setspace}

\usepackage{xcolor}

\begin{document}

\title{Prediction of Time and Distance of Trips Using Explainable Attention-based LSTMs}

\author{
 Ebrahim Balouji \\
  Department of Mechanics and Maritime Sciences \\
   Chalmers University of Technology \\
   Gothenburg, Sweden \\
   \And
 Jonas Sj\"{o}blom \\
  Department of Mechanics and Maritime Sciences \\
   Chalmers University of Technology \\
   Gothenburg, Sweden \\
  \And
 Nikolce Murgovski \\
  Department of Electrical Engineering \\
   Chalmers University of Technology \\
   Gothenburg, Sweden \\
  \And
   Morteza Haghir Chehreghani \\
  Department of Computer Science and Engineering \\
   Chalmers University of Technology \\
   Gothenburg, Sweden \\
}

%
%

%
%

%

\maketitle              
\begin{abstract}

In this paper, we propose machine learning solutions to predict the time of future trips and the possible distance the vehicle will travel. For this prediction task, we develop and investigate four methods. In the first method, we use long short-term memory (LSTM)-based structures specifically designed to handle multi-dimensional historical data of trip time and distances simultaneously. Using it, we predict the future trip time and forecast the distance a vehicle will travel by concatenating the outputs of LSTM networks through fully connected layers. The second method uses attention-based LSTM networks (At-LSTM) to perform the same tasks. The third method utilizes two LSTM networks in parallel, one for forecasting the time of the trip and the other for predicting the distance. The output of each LSTM is then concatenated through fully connected layers. Finally, the last model is based on two parallel At-LSTMs, where similarly, each At-LSTM predicts time and distance separately through fully connected layers. Among the proposed methods, the most advanced one, i.e., parallel At-LSTM, predicts the next trip's distance and time with 3.99 $\%$ error margin where it is 23.89 $\%$  better than LSTM, the first method. We also propose TimeSHAP as an explainability  method for understanding how the networks perform learning and model the sequence of information.
\end{abstract}
\section{Introduction}

\label{sec:introduction}
Trip prediction constitutes an important element for intelligent transport in various ways, e.g., simulation of public transportation, traffic analysis, battery charge planning,  demand and load investigation, and preparation of vehicles for trips \cite{chen2018trip,eberstein2022unified,chidlovskii2015improved,zhou2018orientsts,bieler2022survey}.
It also helps to optimize energy management, save energy,  prolong components' lifetime,  and can also be used to support electric power systems \cite{Akerblom0C20,AkerblomCC23}.
In particular, it yields increasing efficiency and reduces the energy consumption of Electric Vehicles (EVs) and even combustion engines.

Predicting the trip time can help prepare the vehicle for the mission by thermally conditioning the passengers' cabin. This is specifically important for batteries or engines to avoid a cold start. In an electric vehicle, cold start may cause a reduction in the efficiency leading to capacity degradation of batteries
and, subsequently, a reduction in their lifetime
\cite{KIM2021121001,shigemoto2022elucidation,min2020thermal,NELSON2002349}.

Distance prediction can help to estimate how much energy will be needed. Also, it can pinpoint if warming batteries before the trip is needed by trading off energy efficiency, battery lifetime, and power availability. Knowing the amount of power needed for the trip can help to optimize charging policies and avoid fast charging, which may be used to take preventive measures for fast battery degradation \cite{LUNZ2012511, osti_1236824}.
Using the information on both time and distance of trips can open the possibility of using batteries for the power flow in electric grid management \cite{8113559,en5104076} and reduce energy consumption. It can also be used to plan to build charging stations in the cities, especially in megacities.
Observing and predicting the general traveling pattern of vehicle users in a region can help to  estimate the overall statistics of
a transportation system.

Reaching all the accomplishments mentioned above highly relies on the possibility of predicting the accurate and exact time and distance of future travels. However, passengers do not always
travel according to a linear planning schedule, and the trips can be made according to stochastic patterns. This stochasticity may even increase when predicting a fleet of vehicles.
Several works have been proposed to predict the possible destination for a trip/vehicle \cite{panahandeh2017driver,epperlein2018bayesian,ermagun2017real,chen2018trip,eberstein2022unified} and the path/route to be taken under different conditions \cite{Akerblom0C20,AkerblomCC23,AkerblomHC23}. However,  the prediction of the exact time of the trip and the distance that the vehicle will travel are not studied.
In this paper, we develop two approaches to address this task.
We propose predicting future trips' time and distance by analyzing historical data using unique features-specific learning models based on LSTM and attention-based LSTM (At-LSTM) neural networks.
We also develop timeSHAP, a posthoc model-agnostic explainability method specifically tailored for LSTM and At-LSTM handling multidimensional and length-variant sequences of information.

The contributions of this paper can be summarized as follows:
i)  We develop a novel deep learning-based framework utilized by At-LSTM-based structure to predict future trip time and distance that car will travel.
ii)  Using the most advanced method, the prediction error significantly decreases by about $\approx 22 \%$ compared to using only LSTM.
iii) The proposed method has minimal preprocessing steps and is a data-driven approach with a diverse dataset obtained from 700 vehicles.
iv) we also  develop timeSHAP as a model-agnostic LSTM explainer that is utilized based on  KernelSHAP and expands it to analyzing sequential data.
%
%
The rest of the paper is organized as follows: In section 2, we introduce the data, the four different models as well as the explainability component. In section 3, we demonstrate the experiments and discuss the respective results. Finally, we conclude the paper in section 4.

\section{Methodology }
\label{sec:Methodology}

In this section, we describe four different
methods, including two LSTM based and two At-LSTM-based networks, as four solutions to predict the
time of future trips together with the possible distance that will be traveled by the vehicle. 

\begin{figure}
  \begin{center}
    \includegraphics[width=0.5\textwidth]{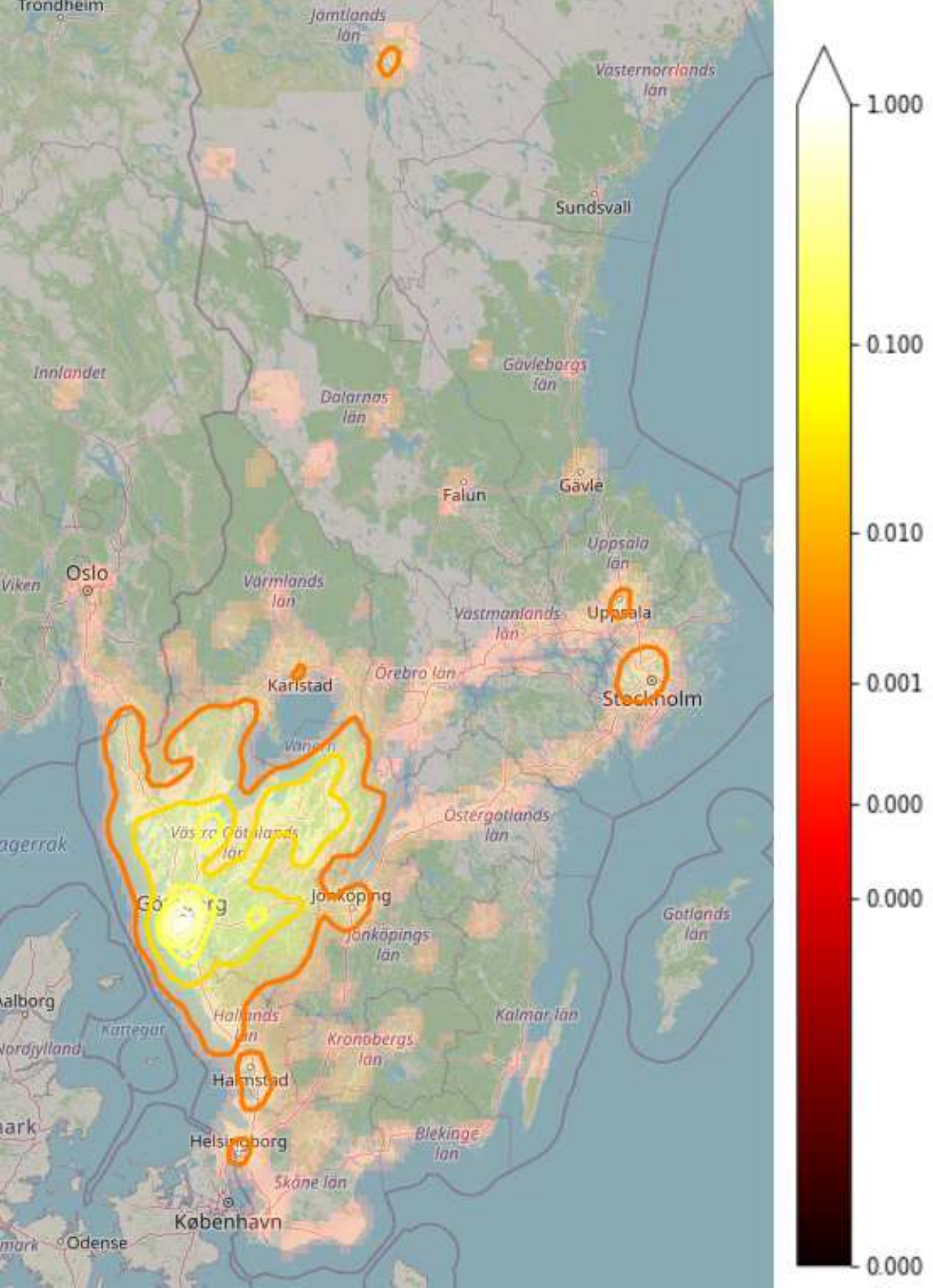}
  \end{center}
  \caption{Density map of the trips made by 700 cars in different cities in Sweden. The figure is adapted from \cite{eberstein2022unified}.}
  \label{fig:density}
\end{figure}

\subsection{Data and preprocessing}
The dataset used in this study is a real-world dataset where
95962 trips have been collected by 700 GPS-tracked vehicles in Sweden. To avoid GPS disconnection, trips made within 10 min  are considered one trip.
Furthermore, the trips shorter than 3 km are discarded since the cold start of the vehicles are more energy efficient for short-distance travels. After applying such preprocessing steps, about $50\%$ of the original trips and $59\%$ of
the vehicles are included in the filtered dataset. The density map of the trips based on GPS coordinates and the destination is illustrated in Fig. \ref{fig:density}.
More detail about the trips can be found in \cite{karlsson2013swedish}.

We define $\Delta t$ as in how many seconds the next trip will be made, i.e.,
\begin{equation}
    \Delta t_{n}= t_n -t_{n-1} ,
\end{equation}
where $t_n$ and $t_{n-1}$ are the time of the last two trips in historical data. We also define $d_n$ as the distance the vehicle has traveled. Finally, to consider the possibility of variation of the trip frequency on different days of the week, we consider the corresponding weekday as a feature alongside $\Delta t$ and $d$.  Therefore prediction of the future trip time and distance  by observing the historical data can be formulated as follows.
\begin{equation}
    \begin{bmatrix}
        \Delta t_{n+1}\\
        d_{n+1}
    \end{bmatrix}\\
    =f{\begin{bmatrix}
\left(\Delta t_{n-1}, \Delta t_{n-2}, \Delta t_{n-3}, \ldots, \Delta t_n\right) \\
\left(d_{n-1}, d_{n-2}, d_{n-3}, \ldots, d_n\right) \\
\text { (corresponding weekday of each trip) }
 \end{bmatrix}},
 \label{eq:features}
\end{equation}
where $f$ represents, for example, a deep learning model.

\begin{figure}
\centering
\subfloat[][]{\includegraphics[width=.53\linewidth]{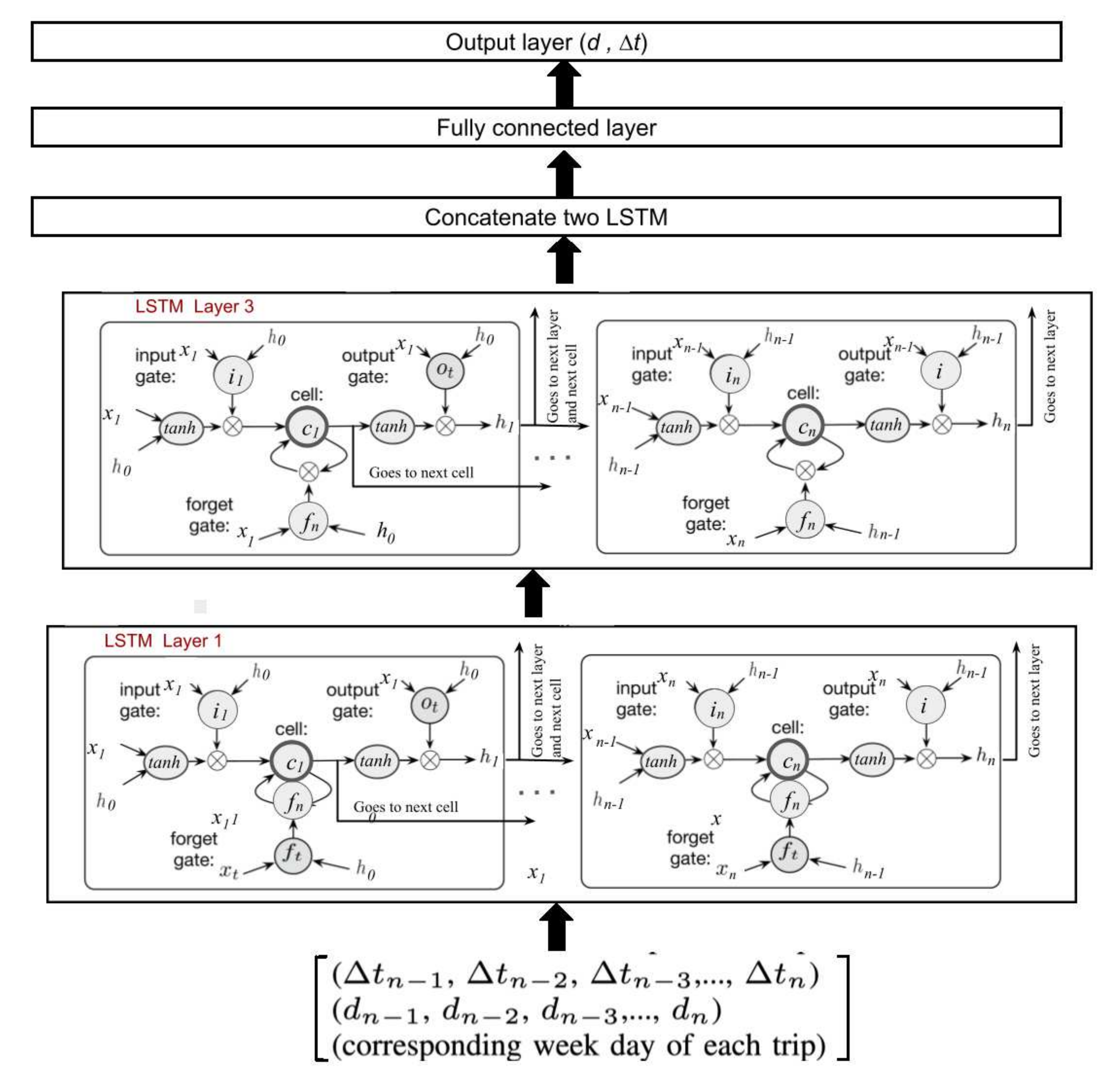}}
\subfloat[][]{\includegraphics[width=.53\linewidth]{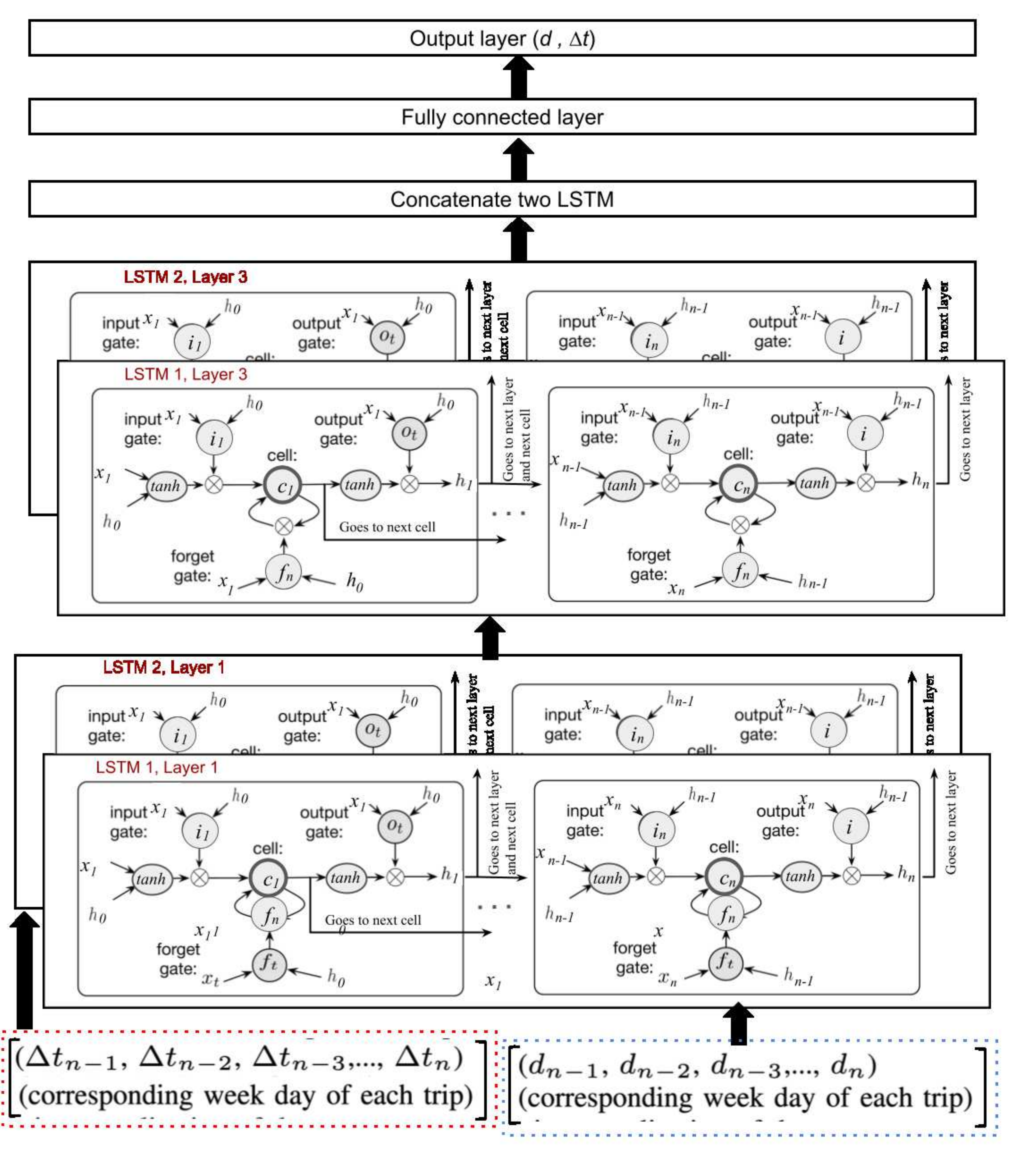}}
\caption{The unrolled diagram of (a) LSTM  and (b) parallel LSTM method.  $x$, $h$, $f$, $i$, $c$, and $o$ stand for input data, state and forget gate, input gate, cell information, and output gate of LSTM, respectively. $W$ and $U$ stand for weights for each gate.
\vspace{3mm}}
\vspace{-5mm}
\label{fig:single_LSTM}
\end{figure}

\subsection{ LSTM based methods}
\begin{figure*}[htb!]
\subfloat[][]{\includegraphics[width=.5\linewidth]{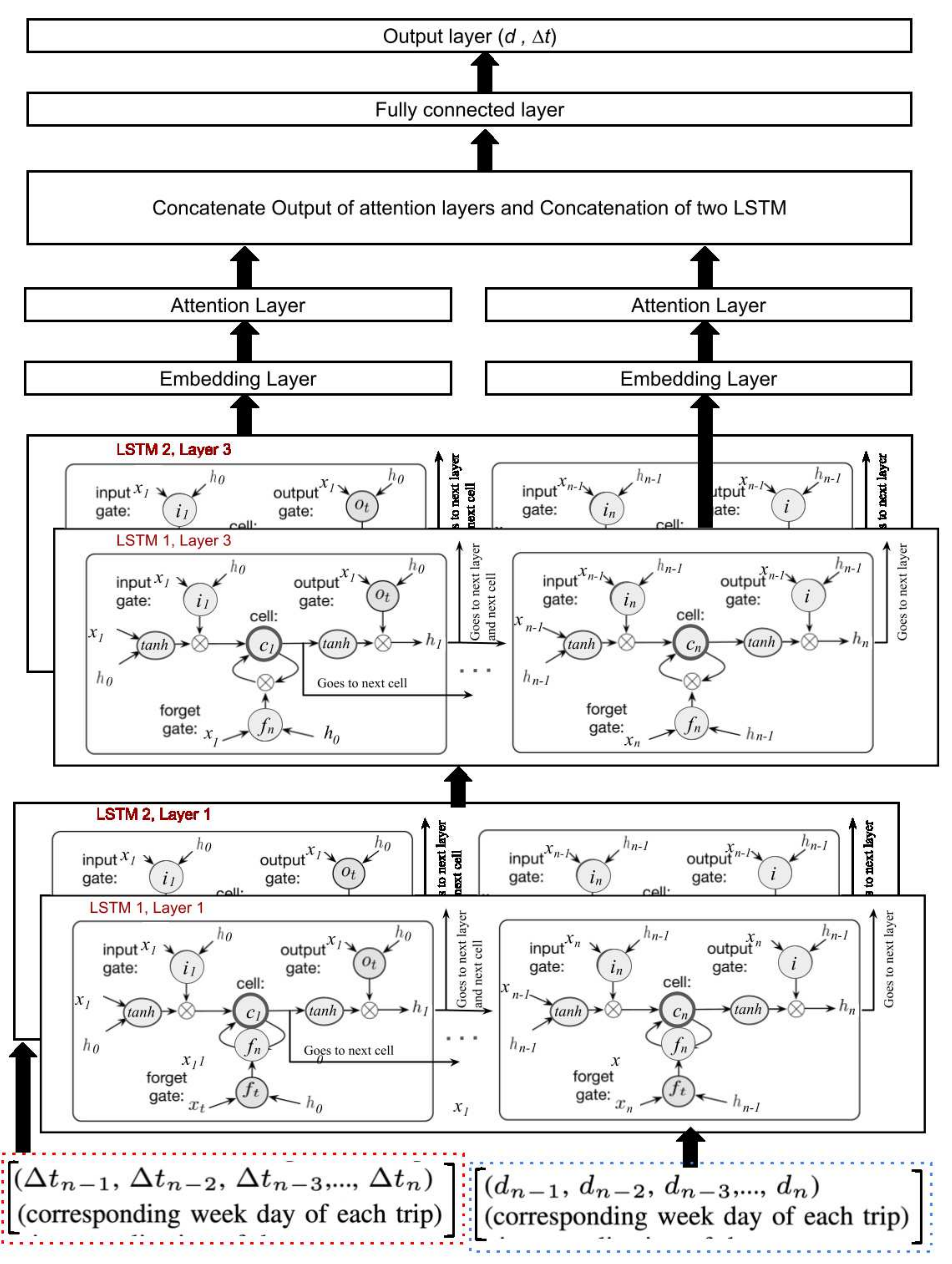}
}
\subfloat[][]
{\includegraphics[width=.5\linewidth]{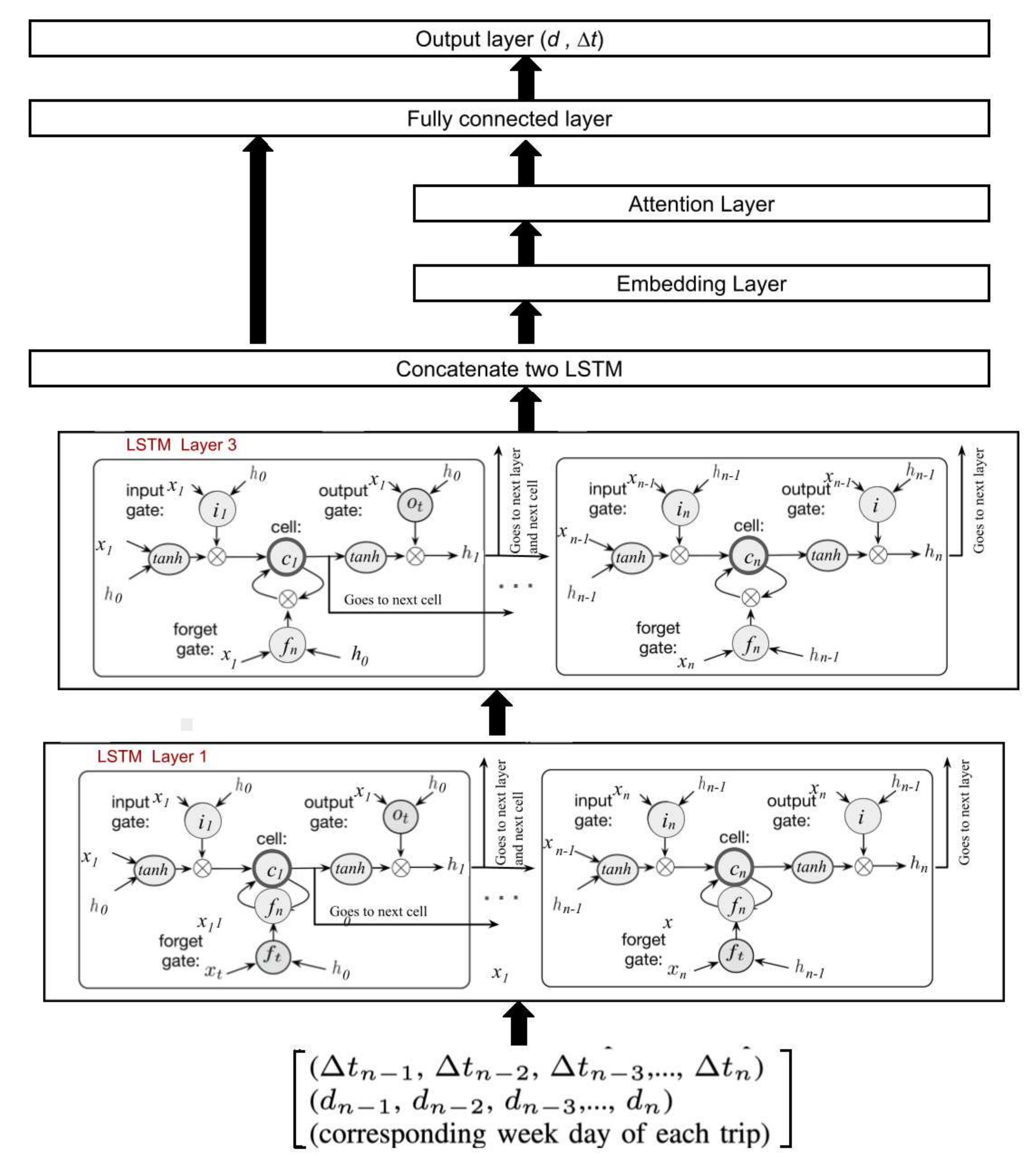}}
\caption{The unrolled diagram of (a) parallel LSTM and (b) Parallel At-method. $x$, $h$, $f$, $i$, $c$, and $o$ stand for input data, state and forget gate, input gate, cell information, and output gate of LSTM. $W$ and $U$ represent the weights. 
\vspace{8mm}
}
\vspace{-8mm}
\label{fig:At-LSTM}
\end{figure*}

The first method (called $PM1$) consists of three LSTM layers followed by two fully connected layers (see Fig.~\ref{fig:single_LSTM} (a)).
The input to the LSTM networks is a multidimensional matrix introduced in Eq. \eqref{eq:features}.
The second method is based on using two parallel LSTM-based neural networks where each branch of the network receives separate $\Delta t$ and $d$, and the outputs of the two LSTMs are then concatenated through two fully connected layers. Fig.~\ref{fig:single_LSTM} (b) illustrates the structure of the second method (called $PM2$). The third method (called $PM3$) is the \emph{attention} version of the LSTM-based model developed in $PM1$.  Furthermore, due to the inequality of the length of the selected sequences of the trips, we use an embedding layer to equalize the length of the state vector for each LSTM layer and feed them to the attention layer. The structure of $PM3$ is shown in Fig. \ref{fig:At-LSTM} (a).
Finally, the last method is the attention-based version of the $PM2$, where two parallel LSTM networks are processing $\Delta t$ and $d$ separately. Furthermore, the outputs of the LSTM layers go to separate attention layers, after which they are embedded using the embedding layer again. Finally, the outputs of the LSTM and  attention layers are fed into a fully connected layer. Fig. \ref{fig:At-LSTM} (b) illustrates the last method, called $PM4$.

In the following, we  describe the fundamental components of these methods, such as LSTM networks, embedding, and attention layer. In addition, we augment the models with an explainability component, as described thereafter.

\subsubsection{LSTM networks}

An LSTM cell, as a stateful operator, is a type of RNN that analyzes the $\Delta t$ and the distance $d$ of the trips  sequentially \cite{40olah2015understanding}. For the sake of simplicity, we  represent $\Delta t$ and $d$, the input vectors to LSTM, together by $x_n$, where $n$ is the length of the sequence. For each $n$,
$x_n$ is used to compute an output $y_n$ and updates the internal state of the cell at each layer which can be defined as:
\begin{gather}
    h_n = f_W(h_{n-1}, x_n)
\end{gather}
where $h_n$ is a new state and $f_W$ is a nonlinear function parameterized by a set of weights $W$.
$h_n$ implicitly encodes the learned representation of the non-linearity and time-variance of the modeled data. For the LSTM cells, the exact update rules for $y_n$ and $h_n$ are defined in (\ref{eq:LSTM_output})-(\ref{eq:LSTM_state}).

\begin{gather}
    y = \begin{bmatrix}
           i \\
           f \\
           o \\
           g
        \end{bmatrix}
    = \begin{bmatrix}
           \sigma \\
           \sigma \\
           \sigma \\
           \tanh
        \end{bmatrix}
    W
    \begin{bmatrix}
           h_{n-1} \\
           x_n
        \end{bmatrix}, \label{eq:LSTM_output}
\end{gather}
\vspace{-9pt}
\begin{gather}
    c_n = f \odot c_{n-1} + i \odot g,
\end{gather}
\vspace{-20pt}
\begin{gather}
    h_n = o \odot \tanh (c_n), \label{eq:LSTM_state}
\end{gather}
where $i$ is the input gate to write information into the cell, $f$ is the \textit{forget gate} that receives the data for resetting the cell state, $g$ determines the magnitude of the influence of the input $x_n$ on the cell and $o$ represents the output gate which controls the information flow to the next layer. $\sigma$ and $\tanh$ are non-linear activation functions necessary for modelling non-linear data and $\odot$ is element-wise multiplication. All these cells, in combination, provide a powerful tool to learn features and model the behavior of time-variant data in order to be used as a predictor. The parallel version of the LSTM method in the mathematical derivation is illustrated in Algorithm (I).

\begin{algorithm}
\scriptsize
    \caption{\textit\scriptsize{LSTM algorithm  }: }
\label{algorithm:}
$n$  \text{: number of trips in selected window of the historical data }\\
$d$  \text{: the distance that a vehicle travels on each trip }\\

$k$  \text{: the number of LSTM networks }\\
$\Delta t$  \text{: time difference of the trip from the previous trip }\\

Read ($\Delta t_{n-1}$, $\Delta t_{n-2}$, $\Delta t_{n-3}$,..., $\Delta t_{n}$) \\
Read ($d_{n-1}$, $d_{n-2}$, $d_{n-3}$,..., $d_{n}$) \\

Read (corresponding weekday of each trip) \\
Max-min normalization of data

\SetKwBlock{DoParallel}{do in parallel}{end}

\DoParallel{

$k \leftarrow k+1 $\\

\textbf{LSTM Networks 1 (for prediction of $\Delta t_{n+1}$)}\\
\textbf{LSTM Networks 2 (for prediction of $d_{n+1}$)}\\

\While{$i<=n$ ( number of neurons in $FC^1$)}{
\textbf{First LSTM layer:}\\
$n \leftarrow n+1 $\\
$i_{n}^{1}=\sigma_{g}\left(W_{i} F C^{1,k}+U_{i} h_{n-1}^{1,k}+b_{i}\right) \quad \triangleright $ input gate\\
$f_{n}^{1,k}=\sigma_{g}\left(W_{f} F C^{1,k}+U_{f} h_{n-1}^{1,k}+b_{f}\right) \quad \triangleright$ forget gate\\
$o_{n}^{1,k}=\sigma_{g}\left(W_{o} F C^{1,k}+U_{o} h_{n-1}^{1,k}+b_{o}\right) \quad \triangleright$ output gate\\
$\tilde{c}_{n}^{1,k}=\tanh \left(W_{c} F C^{1,k}+U_{c} h_{n-1}^{1,k}+b_{c}\right)$\\
$c_{n}^{1}=f_{n}^{1} \circ c_{n-1}^{1}+i_{n}^{1,k} \circ \tilde{c}_{n}^{1,k} \quad \triangleright$ cell information \\
$h_{n}^{1,k}=o_{n}^{1,k} \circ \tanh \left(c_{n}^{1,k}\right) \quad \triangleright$ state goes to next layer

$\bullet$\\
$\bullet$\\
$\bullet$\\

 $\mathbf{3,k}^\textbf{rd}$
 \textbf{LSTM layer:}\\
$i_{n}^{3,k}=\sigma_{g}\left(U_{i} h_{n-1}^{3,k}+b_{i}\right) \quad \triangleright$ input gate\\
$f_{n}^{3,k}=\sigma_{g}\left(U_{f} h_{n-1}^{3,k}+b_{f}\right) \quad \triangleright$ forget gate\\
$o_{n}^{3,k}=\sigma_{g}\left(U_{o} h_{n-1}^{3,k}+b_{o}\right) \quad \triangleright$ output gate\\
$\tilde{c}_{n}^{3,k}=\tanh \left(U_{c} h_{n-1}^{3,k}+b_{c}\right)\quad$\\
$c_{n}^{3,k}=f_{n}^{3,k} \circ c_{n-1}^{3,k}+i_{n}^{3,k} \circ \tilde{c}_{n}^{3,k} \quad \triangleright$ cell information\\
$h_{n}^{3,k}=o_{n}^{3,k} \circ \tanh \left(c_{n}^{3,k}\right) \quad \triangleright$ state goes to FC layer

 }}
  \textbf{Concatenate the result of two LSTM networks :}\\
\begin{align*}
 CL=Concatenate. \begin{bmatrix}
h_1^{3,1}& h_2^{3,1}&...&h_n^{3,1}\\h_1^{3,2}& h_2^{3,2}&...&h_n^{3,2}
\end{bmatrix}\\
\end{align*}

 \textbf{First FC layer:}\\
$FC^1=W_{F}.(CL)$\\
\textbf{Second FC layer:}\\
 $[\Delta t_{n+1} , d_{n+1}] = W_{F}.(FC^1)$

\end{algorithm}

\subsubsection{Attention-based version of LSTM}

The standard LSTM cannot detect which part of the  sequence is relevant  for aspect-level forecasting.
In order to address this issue, we design an attention-based mechanism that can capture the
key part of historical data in response to a given aspect.
Fig. \ref{fig:At-LSTM} (a) represents the architecture of this At-LSTM model.

Let $H \in \mathbb{R}^{k \times n}$
be a matrix consisting of hidden vectors $[h_1, . . . , h_n ]$ that the LSTM produces,
where $k$ is the size of hidden layers and $n$ is the
length of the given sentence. Furthermore, $v_a$ represents the embedding of  both the aspect and $e_N \in \mathbb{R}^N$
vectors. The attention mechanism produces
an attention weight vector $\alpha$ and a weighted hidden representation $r$, as follows.

\begin{equation}
M=\tanh \left(\left[\begin{array}{c}
W_h H \\
W_v v_a \otimes e_N
\end{array}\right]\right),
\label{eq:qe}
\end{equation}

\begin{equation}
\alpha=\operatorname{sigmoid}\left(w^T M\right),
\end{equation}

\begin{equation}
r=H \alpha^T,
\end{equation}

where $M \in \mathbb{R}^{\left(k+k_a\right) \times n}, \alpha \in \mathbb{R}^N, r \in \mathbb{R}^k$ $W_h \in \mathbb{R}^{k \times k}, W_v \in \mathbb{R}^{k_a \times k_a}$ and $w \in \mathbb{R}^{k+k_a}$ are
projection parameters. $\alpha$ is a vector consisting of
attention weights and $r$ is a weighted representation
of $h$ produced by LSTMs. The operator $\otimes$ in $v_a \otimes e_N=[v ; v ; \ldots ; v]$ means that the operator repeatedly concatenates $v$ for $n$ times, where
$e_n$ is a column vector with $n$ samples. $W_v v_a \otimes e_n$ is repeating the linearly transformed $v_a$ as many times
as there are data points in the selected window of data.
The last state representation can be defined as:
\begin{equation}
\centering
h^\star = \tanh(W_{p^r} + W_xh_n ),
\end{equation}
where, $h^\star$ $\in$ $R_k, W_p $ and $W_x$ are the weights to be inferred during training.  Motivated
from \cite{rocktaschel2015reasoning}, we find that this model
works practically better if we add $W_x h_n$ into the final representation of the sequence of information.

The attention mechanism helps the model to
apprehend the  essential part of the information in the selected window of historical data where diverse characteristics are assessed.The parallel version of the LSTM method in the mathematical derivation is illustrated in Algorithm (II)

\begin{algorithm}
\scriptsize
    \caption{\textit\scriptsize{LSTM algorithm  }: }
\label{algorithm:}
$n$  \text{: number of trips in selected window of the historical data }\\
$d$  \text{: the distance that a vehicle travels on each trip }\\

$k$  \text{: the number of LSTM networks }\\
$\Delta t$  \text{: time difference of the trip from the previous trip }\\

Read ($\Delta t_{n-1}$, $\Delta t_{n-2}$, $\Delta t_{n-3}$,..., $\Delta t_{n}$) \\
Read ($d_{n-1}$, $d_{n-2}$, $d_{n-3}$,..., $d_{n}$) \\

Read (corresponding weekday of each trip) \\
Max-min normalization of data

\SetKwBlock{DoParallel}{do in parallel}{end}

\DoParallel{

$k \leftarrow k+1 $\\

\textbf{LSTM Networks 1 (for prediction of $\Delta t_{n+1}$)}\\
\textbf{LSTM Networks 2 (for prediction of $d_{n+1}$)}\\

\While{$i<=n$ ( number of neurons in $FC^1$)}{
\textbf{First LSTM layer:}\\
$n \leftarrow n+1 $\\
$i_{n}^{1}=\sigma_{g}\left(W_{i} F C^{1,k}+U_{i} h_{n-1}^{1,k}+b_{i}\right) \quad \triangleright $ input gate\\
$f_{n}^{1,k}=\sigma_{g}\left(W_{f} F C^{1,k}+U_{f} h_{n-1}^{1,k}+b_{f}\right) \quad \triangleright$ forget gate\\
$o_{n}^{1,k}=\sigma_{g}\left(W_{o} F C^{1,k}+U_{o} h_{n-1}^{1,k}+b_{o}\right) \quad \triangleright$ output gate\\
$\tilde{c}_{n}^{1,k}=\tanh \left(W_{c} F C^{1,k}+U_{c} h_{n-1}^{1,k}+b_{c}\right)$\\
$c_{n}^{1}=f_{n}^{1} \circ c_{n-1}^{1}+i_{n}^{1,k} \circ \tilde{c}_{n}^{1,k} \quad \triangleright$ cell information\\
$h_{n}^{1,k}=o_{n}^{1,k} \circ \tanh \left(c_{n}^{1,k}\right) \quad \triangleright$ state goes to next layer

$\bullet$\\
$\bullet$\\
$\bullet$\\

 $\mathbf{3,k}^\textbf{rd}$  \textbf{LSTM layer:}\\
$i_{n}^{3,k}=\sigma_{g}\left(U_{i} h_{n-1}^{3,k}+b_{i}\right) \quad \triangleright$ input gate\\

$f_{n}^{3,k}=\sigma_{g}\left(U_{f} h_{n-1}^{3,k}+b_{f}\right) \quad \triangleright$ forget gate\\
$o_{n}^{3,k}=\sigma_{g}\left(U_{o} h_{n-1}^{3,k}+b_{o}\right) \quad \triangleright$ output gate\\
$\tilde{c}_{n}^{3,k}=\tanh \left(U_{c} h_{n-1}^{3,k}+b_{c}\right)\quad$\\
$c_{n}^{3,k}=f_{n}^{3,k} \circ c_{n-1}^{3,k}+i_{n}^{3,k} \circ \tilde{c}_{n}^{3,k} \quad \triangleright$ cell information\\
$h_{n}^{3,k}=o_{n}^{3,k} \circ \tanh \left(c_{n}^{3,k}\right) \quad $

\begin{align*}
    V_a=embed\begin{bmatrix}
h_1^{1,k}& h_2^{1,k}&...&h_n^{1,k}\\h_1^{2,k}& h_2^{2,k}&...&h_n^{2,k}
\\h_1^{3,k}& h_2^{3,k}&...&h_n^{3,k}
\end{bmatrix} \\
\end{align*}

 $\alpha=sigmoid.(W_{F}.$ $v_a)$ $\triangleright$ state goes to concatenation layer\\} }

\textbf{Concatenate the result of two LSTM networks :}\\

\begin{flalign*}
CL=Concatenate.
\begin{bmatrix}
Attention^1\\
Attention^2\\
h_1^{3,1}& h_2^{3,1}&...&h_n^{3,1}\\h_1^{3,2}& h_2^{3,2}&...&h_n^{3,2}
\end{bmatrix} \\
\end{flalign*}

 \textbf{First FC layer:}\\
$FC^1=Relu (W_{F}.(CL))$\\
\textbf{Second FC layer:}\\
 $[\Delta t_{n+1} , d_{n+1}] = sigmoid(W_{F}.(FC^1))$

\end{algorithm}

\subsection{Explainability}
Here, we develop an explainer that is not only
model-agnostic and post-hoc, but also specifically tailored to the At-LSTM networks.
To be able to explain the predictor algorithm, the method must provide both
trip and  attributions of extracted features throughout the selected window of the sequence from historical data. The aim is to provide  explainability of  the sequential analysis while maintaining three desirable properties:  accuracy, missingness, and consistency \cite{shapley1953value}.
Therefore, we use an explainable, model-agnostic At-LSTM
that is developed based on the TimeSHAP method introduced in \cite{bento2021timeshap}. TimeSHAP defines a \emph{significance} index to the features or trips  in the input that specifies the extent to which those
features or trips impact the prediction task. To describe a piece of sequential information, $X \in R^{fxD}$, with $D$ trips and $f$ features of each trip, TimeSHAP fits a linear function $g$ (explainer) that matches the local
output of a complex explainer $g$ by minimizing the loss function defined in \cite{bento2021timeshap}.

The function $g$ that approximates $f\left(h_X(z)\right)$ is denied as
\begin{equation}
f\left(h_X(z)\right) \approx g(z)=b_0+\sum_{i=1}^m w_i z_i,
\end{equation}
where the term $b_0=f\left(h_X(\mathbf{0})\right)$ is a bias value and resembles the output model with trips and features ticked off (dubbed base score). The set $\{w_i, i \in\{1, \ldots, m\}\}$ refers to the weights representing the impact of each trip/feature ($m = f$ or $m = D$), for each dimension that is going to be
explained. The perturbation function $h_X:\{0,1\}^m \mapsto \mathbb{R}^{d \times l}$  maps a coalition $z \in\{0,1\}^m$  to the original input space $x \in R^{f\times D}$. Note that
the sum of significance indices for all the features corresponds to the difference
between the score of the model   $f(X)=f\left(h_X(\mathbf{1})\right)$ and the score  which is $h_X(\mathbf{0})$.

\section{Experimental Results}

In this section, we investigate the different models and compare their performances on our real-world dataset.
We perform grid-search-based hyperparameter tuning for all methods to set their hyperparameters.

Considering the training time for each hyperparameter choice, we have selected 25 $\%$ of data divided into test and training parts. The hyperparameters are aimed tuned, where the range of the search, steps on each search, and the corresponding results are shown in Table \ref{table:hyper}. As can be seen, the choice of the window of data, LSTM structure in all methods, and learning rate have the highest variation effect on the prediction error.

\begin{table}
\scriptsize
\caption{Investigation of the effect of the hyperparameter tuning and their results for the proposed methods. Abbreviations used on this table: parameters (P), methods(M), error (err.), variation (Var.), average (Avg.) learning rate (lr),  Mean Absolute Error (MAE), Mean Squared Error (MSE), Log Hyperbolic Cosine (LHC), Huber loss (HL), Mean Squared Logarithmic Error (MSLE), Poisson loss (PS) }

\label{table:hyper}
\renewcommand{\arraystretch}{1.3}
\small
\centering
\scriptsize

\begin{tabular}{lllccc}
\toprule
\toprule
\backslashbox{\textbf{M}}{\textbf{P}} &\multirow{1}{12em}{\textbf{hyperparameters}}&\multirow{1}{6em}{\textbf{best value}}&\multirow{1}{9em}{\textbf{P.err. Var.($\%$)}}
\\
\hline
\multirow{3}{6em}{\textbf{LSTM}} & window size (1:1:14 days) & 5 & 23.8  \\
 & LSTM layers (1:1:5)& 3& 18\\
& $\#$ neurons in LSTM layers (20:10:150)& 60,120,60 & 19  \\
& FL layers (1:1:3) & 2&3  \\
&batch size (16:16:512) & 128 & 5.2 \\
& lr (0.00001:0.05:0.1) & 0.01 &17  \\
& optimizer (SGD, Adam, Adagrad, RMSProp) & Adam &  6.5 \\
& Loss (MAE, MSE, LHC, HL, MSLE, PS) & MAE &  7.3 \\

\hline
\multirow{3}{6em}{\textbf{Parallel LSTM}} & window size (1:1:14 days) & 8 & 25.8  \\
 & LSTM layers (1:1:5)& 3& 15\\
& $\#$ neurons in LSTM layers (20:10:150)& 40,60,40 & 10  \\
& FL layers (1:1:3) & 2& 3\\
&batch size (16:16:512) & 128 &6\\
& lr (0.00001:0.05:0.1) & 0.01& 19\\
& optimizer (SGD, Adam, Adagrad, RMSProp) & Adam &  5.3 \\
& Loss (MAE, MSE, LHC, HL, MSLE, PS) & MAE &  10.3 \\

\hline
\multirow{3}{6em}{\textbf{Attention based LSTM}} & window size (1:1:14 days) & 8 & 25.8  \\
 & LSTM layers (1:1:5)& 3& 15\\
& $\#$ neurons in LSTM layers (20:10:150)& 60,90,60 & 23  \\
& $\#$ neurons in LSTM attention (4:4:256)& 64 & 21  \\

& FL layers (1:1:3) & 2& 3\\
&batch size (16:16:512) & 128 &5\\
& lr (0.00001:0.05:0.1) & 0.01&17\\
& optimizer (SGD, Adam, Adagrad, RMSProp) & Adam &  6.2 \\
& Loss (MAE, MSE, LHC, HL, MSLE, PS) & MAE &  12.3 \\

\hline
\multirow{3}{6em}{\textbf{Parallel Attention based LSTM}} & window size (1:1:14 days) & 8 & 25.8  \\
 & LSTM layers (1:1:5)& 3& 15\\
& $\#$ neurons in LSTM layers (20:10:150)& 40,60,40 & 10  \\
& $\#$ neurons in LSTM attention (4:4:256)& 64 & 10  \\

& FL layers (1:1:3) & 2& 3\\
&batch size (16:16:512) & 128 &5\\
& lr (0.00001:0.05:0.1) & 0.01&17\\
& optimizer (SGD, Adam, Adagrad, RMSProp) & Adam &  6.2 \\
& Loss (MAE, MSE, LHC, HL, MSLE, PS) & MAE &  12.3 \\
\bottomrule
\end{tabular}
\label{tb:hp}
\end{table}

\begin{table}
 \caption{Investigation of proposed methods and their results for the two training fashions.  Abbreviations used on this table: prediction (Pred.), error (Err.), validation (Val.)}
\label{table:results}
\renewcommand{\arraystretch}{1.3}
\small
\centering
\scriptsize
\begin{tabular}{c|c|c|c|c}
\toprule
\toprule
\textbf{Scenario $\backslash$ Methods}&{\textbf{single LSTM}} & {\textbf{\textbf{Parallel LSTM}}}&{\textbf{\textbf{ At-LSTM }}}&{\textbf{Parallel At-LSTM }}\\
\hline
\textbf{Pred. Err. ($\%$) for } & 26.91 &  12.76&13.32 & 4.31 \\
\textbf{ 70 ($\%$) Training 15 ($\%$) Val.} &  &   & \\
\textbf{  15 ($\%$) Test} &  &   & \\
\hline
\textbf{\textbf{ Pred. Err. ($\%$) for Cross Val.  }} & 26.01 & 10.97 &12.49& 3.99\\
\bottomrule
\bottomrule
\end{tabular}
\end{table}

Note that the evaluation of the performance of the proposed methods is based on the prediction error we obtained using, defined as:
\begin{equation}
 100 \cdot \sqrt{\frac{\sum_{n=1}^N \left| X[n] - \hat X[n]
    \right|^2}{\sum_{n=1}^N \left| X[n] \right|^2}
  },
\label{eq:error}
\end{equation}
 where $X[n]$ is the true value (label) and $\hat X[n]$ is the predicted value.

We perform our experiments \textit{Python} environment using a workstation with an Intel i7 3.40 GHz CPU, 48GB RAM, and two NVIDIA GeForce RTX 4080 GPUs.

\subsection{Comparison of different methods}

\begin{figure*}[thb!]
  \centering
  \subfloat[][]{\includegraphics[trim={0 0 0 0},clip,width=.5\linewidth]{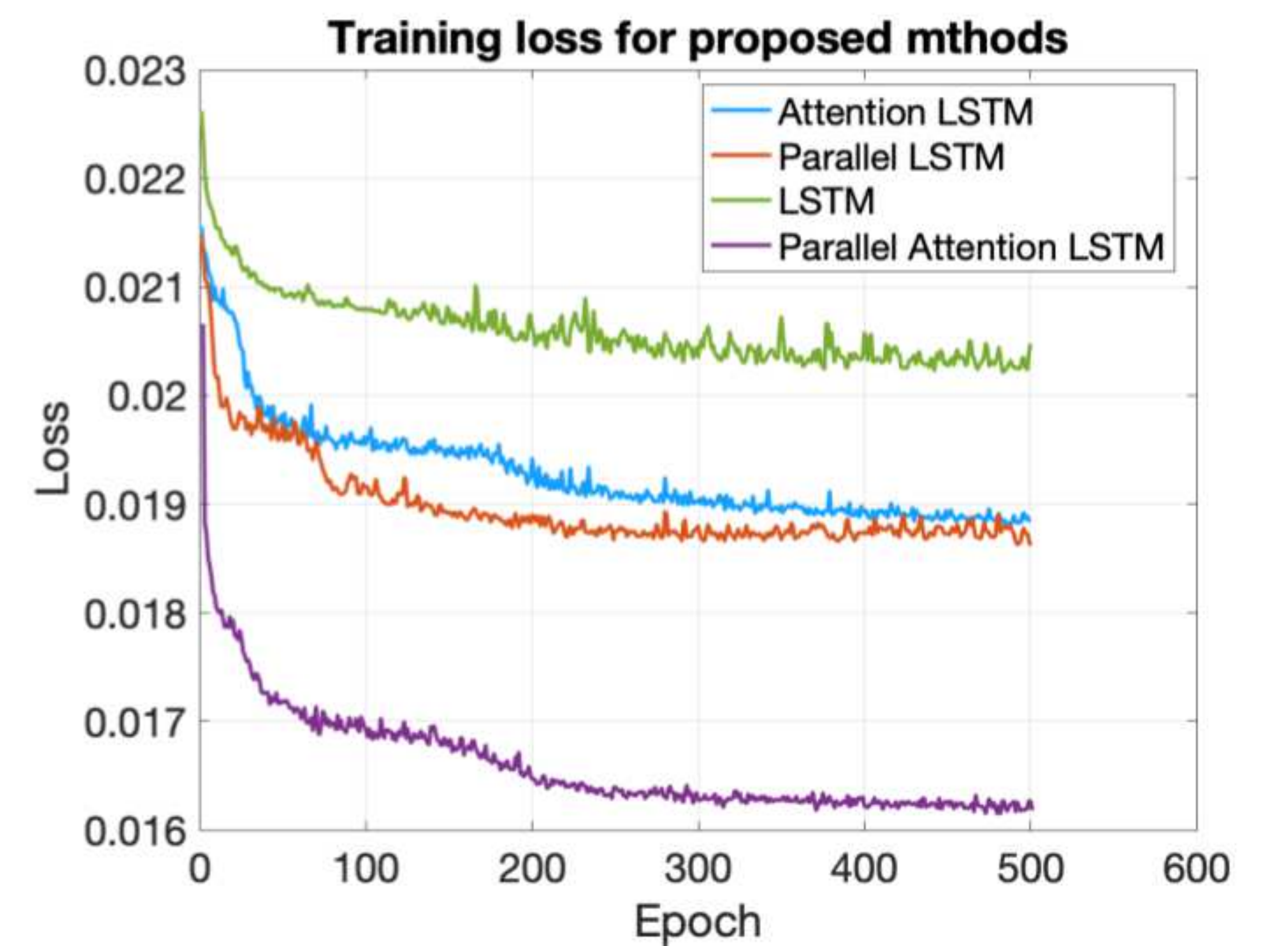}}
  \subfloat[][]{\includegraphics[trim={0 0 0 0},clip,width=.5\linewidth]{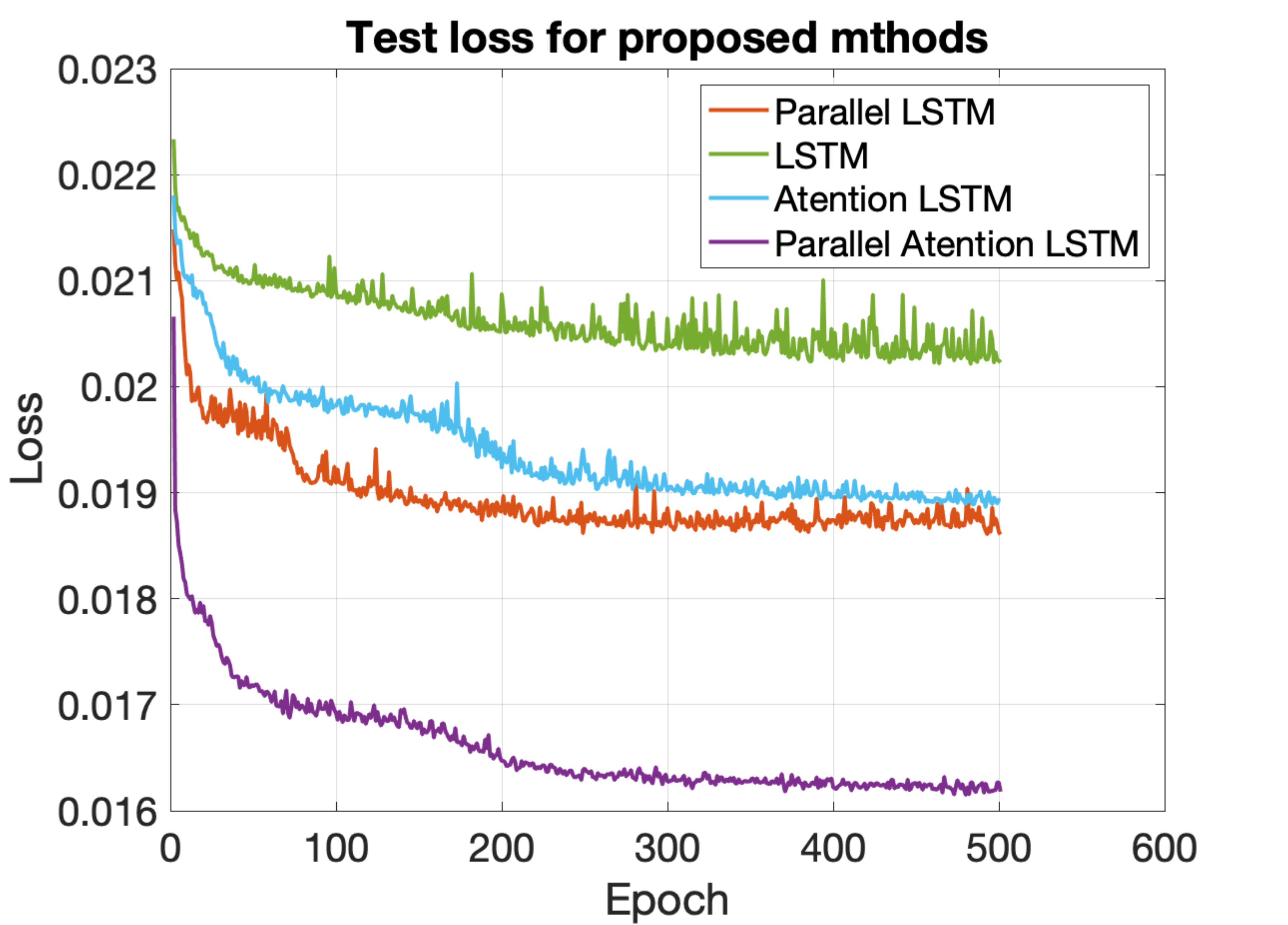}}

  \scriptsize\caption{prediction error on the (a) training phase and (b) test data monitored during the learning process}
 \label{fig:error}
\end{figure*}

\begin{figure*}
  \includegraphics[width=1\linewidth]{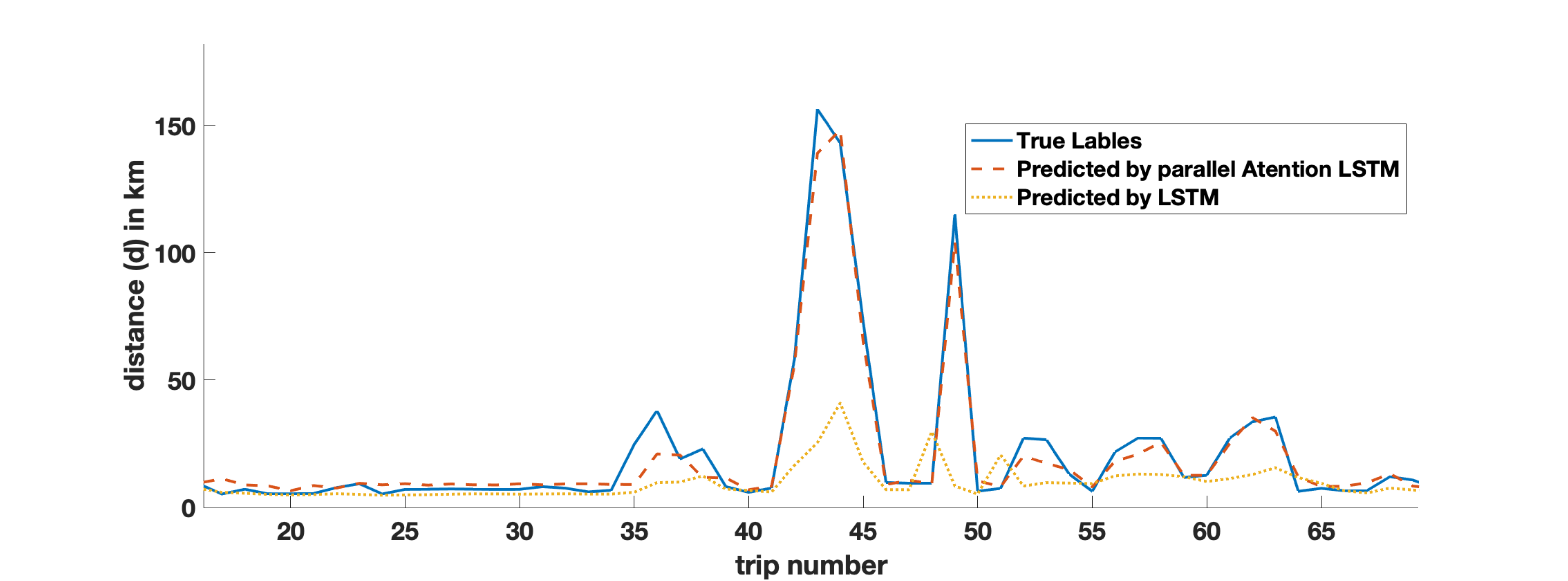}
  \scriptsize\caption{Prediction of the distance of the trips on the part of the test dataset. The blue line is the true labels, the red dashed line is the prediction using parallel At-LSTM, and the yellow dotted line is the prediction of true labels by the LSTM method. }
 \label{fig:perform}

\end{figure*}

Once all the hyperparameters are tuned, we train the networks in two ways. In the first scenario, we divide the data into three subsets: 70$\%$ for training, 15\% for validation, and 15\% for testing. In the second scenario, rather than having a fixed dataset for training and validation, we use cross-validation to train the neural network on each method. We randomly choose 75 $\%$ of the data for training and 15 $\%$ for 10 times. At each training, the weights of the previously trained networks will be updated when seeing a new dataset. Note that the test dataset is never used in the cross-validation process and is the same dataset used as test data for the first case scenario.
We observe that cross-validation-based transfer learning improved the prediction accuracy for all methods (see Table \ref{table:results}). The convergence of the learning procedure is illustrated in Fig.
\ref{fig:error} indicates the prediction error on the training and test dataset per epoch. As an example of the performance of the proposed methods, we illustrate the prediction of the distance of the trips on the part of the test dataset in Fig. \ref{fig:perform}. To avoid cluttering the results, only the LSTM and parallel At-LSTM method predictions are illustrated where they yield the highest and lowest prediction error, respectively. The blue line is the true labels, the red dashed line is the prediction using parallel At-LSTM, and the yellow dotted line is the prediction of true labels by the LSTM method.
Observing the prediction results in Fig.~\ref{fig:perform},  we conclude that parallel At-LSTM indeed performs better in the case of seldom trip patterns, whereas, in the repetitive patterns, pure LSTM yields a similar quality.  Thus, this can be why the prediction accuracy of parallel At-LSTM is higher than the other methods developed in this paper.

\subsection{Explainability}
\begin{figure*}
  \centering
  \subfloat[][]{\includegraphics[trim={0 0 0 0},clip,width=1.\linewidth]{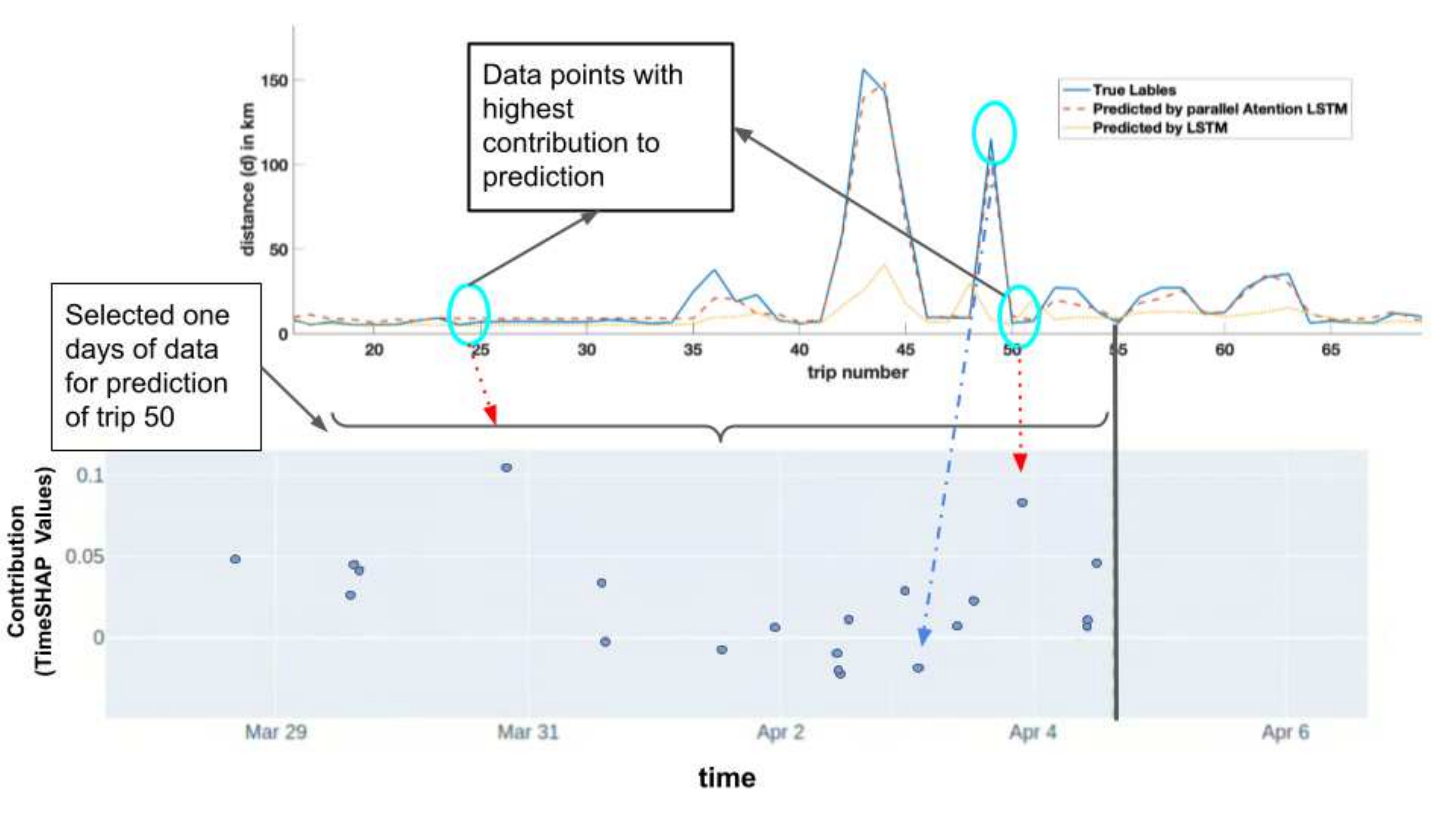}}

  \subfloat[][]{\includegraphics[trim={0 0 0 0},clip,width=1.\linewidth]{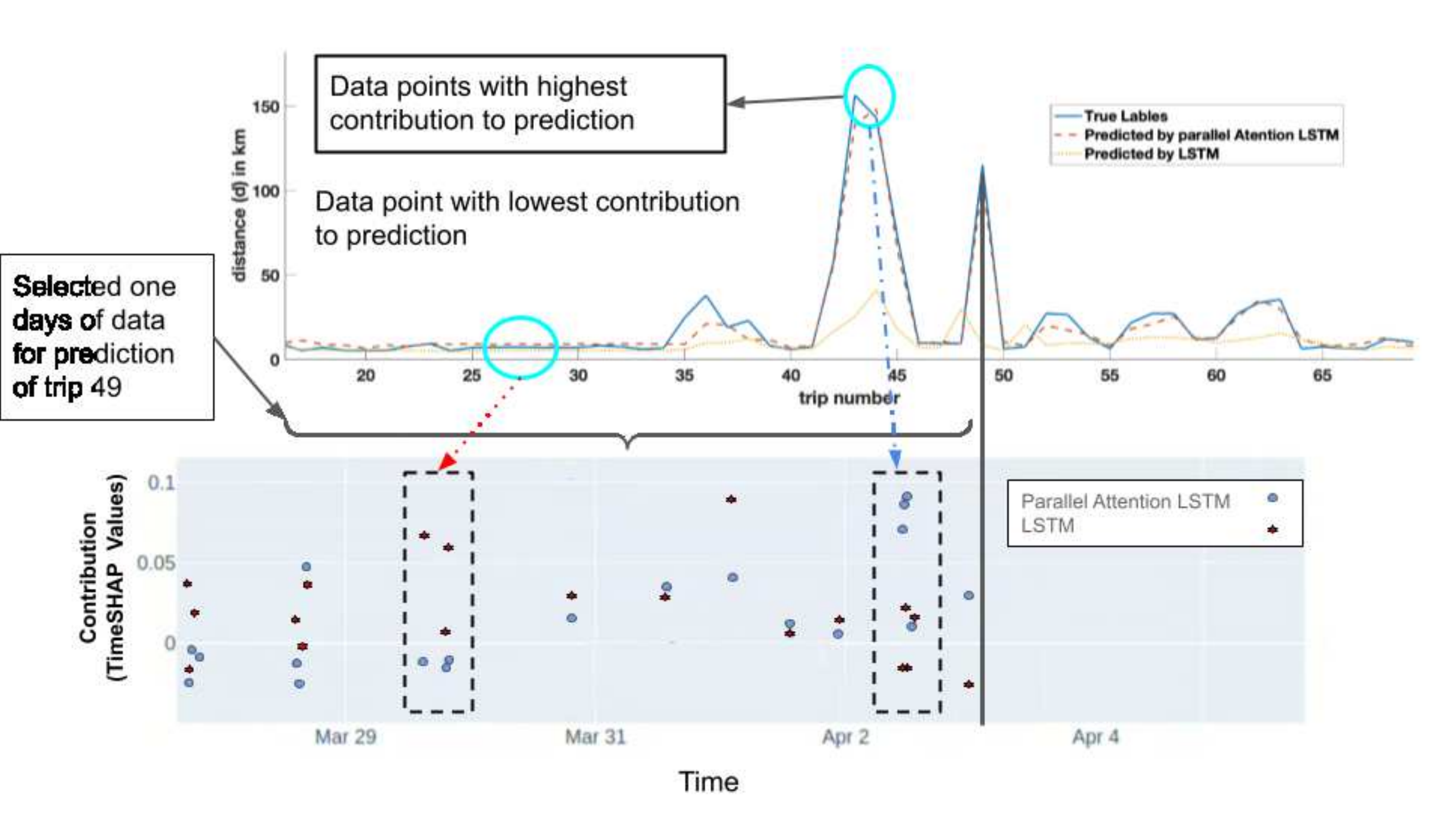}}

  \scriptsize\caption{The Estimated timeSHAP values for the prediction of (a) trip 50 and (b) trip 49 in Fig.~\ref{fig:perform} using seven days of data.}
 \label{fig:explain}
\end{figure*}

As described before, timeSHAP provides local explanations for the model’s behavior regarding a prediction. These descriptions can help improve the model in several aspects, such as auditing or model refining. Nevertheless, end-users, the fraud analysts, can mainly utilize the timeSHAP results to assist their decision-making. It can also be used to gather information and KPIs from end users to develop more sophisticated methods, such as online learning. The result of using the TimeSHAP method for the explainability of the performance of the networks can be seen in Fig. \ref{fig:explain}.
We select two sequences: (i) we observe the trip $d$ from 1-49 and predict the trip number 50, and (ii) we observe trips 5-54 and predict the tip number 55. We extract timeSHAP values for historical trips in both  scenarios. Doing such, indicates the contribution of each trip in the prediction task. Therefore, we can observe the performance of the proposed methods on how well they have learned the relevant patterns in the dataset for an accurate or inaccurate prediction. Considering the fact that the output of all methods is almost the same for scenario (i), to avoid cluttering the results, we only illustrate the timeSHAP values  for the parallel At-LSTM method (see Fig. \ref{fig:explain} (a)). As can be seen from the prediction of the distance of trip number 55, which has a true distance of 5.3 km, the information of trips 51 and 19 (pointed out with red-dotted arrows) have the highest timeSHAP values.  Whereas the trips with distances above 50 km have negative timeSHAP value (see blue dot-dashed arrow). However, in scenario (ii), the timeSHAP values differ as prediction accuracy is significantly low for the LSTM method compared to parallel At-LSTM. Observing the timeSHAP values for both methods in Fig. \ref{fig:explain} (b), the trips around 42-45, which are the longest in the distance, have the highest contribution to the prediction of trip 49. On the other hand, the trips with low distances have the least contribution to the prediction.
On the other hand,  observing the extracted timeSHAP values for the LSTM method in scenario (ii), the trips with the highest distance have a low contribution, and trips with a low distance have the highest contribution. This can be a reason why the prediction accuracy of LSTM is lower compared to the parallel At-LSTM method.

Thereby, we observe that the explainability method is able to successfully suggest the reasoning  behind an accurate prediction. Furthermore, we can also see that the parallel At-LSTM has learned the historical patterns for predicting the future trip distance ($d$). Finally, it may be concluded that the statement "the standard LSTM cannot detect which is the sequence part for aspect-level sentiment forecasting" is true in this setup.

\section{CONCLUSION}\label{sec:conclusion}
In this paper, we proposed a novel deep learning framework for predicting the future  distance and time of vehicle trips. We showed that the parallel attention-based LSTMs yield minimal prediction error compared to using only  LSTM, attention-based LSTM, and two parallel LSTMs.
This method outperformed the alternatives in terms of robustness and learning complicated patterns.   Moreover,
we developed a timeSHAP-based explainability method to show how the patterns have been learned by a  method.
Also, we observed that tuning the parameters and hyperparameters, such as choosing the number of days,  features for creating  sequences, or the learning rate of the optimization algorithms, can affect the prediction accuracy. Thus, it can be concluded that for an accurate predictor model, four important steps need to be followed: the art of feature selection, feature engineering,  innovative selection and structuring  of neural networks, and hyperparameter optimization of the machine learning methods.
In future work, we are planning to create vehicle-specific models where the result of such can create a mother and universal model by using federated learning approaches.

\bibliographystyle{plain}
\bibliography{main}

\end{document}